\documentclass[conference]{IEEEtran}
\usepackage[font=footnotesize]{caption}
\IEEEoverridecommandlockouts
\usepackage{cite}
\usepackage{amsmath,amssymb,amsfonts}
\usepackage{algorithm}
\usepackage{algorithmicx}
\usepackage{algpseudocode}
\usepackage{graphicx}
\usepackage{textcomp}
\usepackage{xcolor}
\usepackage{booktabs}
\usepackage[
    detect-all=true,
    detect-family=true,
    group-digits=false,
    separate-uncertainty=true
]{siunitx}
\usepackage{cleveref}
\usepackage{multirow}
\usepackage{adjustbox}
\usepackage{tikz}
\usepackage{bm}
\usepackage{subcaption}
\usepackage[letterpaper]{geometry}
\def\BibTeX{{\rm B\kern-.05em{\sc i\kern-.025em b}\kern-.08em
    T\kern-.1667em\lower.7ex\hbox{E}\kern-.125emX}}


\renewrobustcmd{\bfseries}{\fontseries{b}\selectfont}
\renewrobustcmd{\boldmath}{}
\newrobustcmd{\B}{\bfseries}
    
\begin{document}
\newgeometry{top=1in, bottom=0.75in, left=0.75in, right=0.75in}
\title{Robust Planning for Autonomous Vehicles with Diffusion-Based Failure Samplers}

\author{\IEEEauthorblockN{Juanran Wang}
\IEEEauthorblockA{\textit{Computer Science} \\
\textit{Stanford University}\\
Stanford, CA, USA \\
jun2026@stanford.edu}
\and
\IEEEauthorblockN{Marc R. Schlichting}
\IEEEauthorblockA{\textit{Aeronautics and Astronautics} \\
\textit{Stanford University}\\
Stanford, CA, USA \\
mschl@stanford.edu}
\and
\IEEEauthorblockN{Mykel J. Kochenderfer}
\IEEEauthorblockA{\textit{Aeronautics and Astronautics} \\
\textit{Stanford University}\\
Stanford, CA, USA \\
mykel@stanford.edu}
}

\maketitle

\begin{abstract}
High-risk traffic zones such as intersections are a major cause of collisions. This study leverages deep generative models to enhance the safety of autonomous vehicles in an intersection context. We train a 1000-step denoising diffusion probabilistic model to generate collision-causing sensor noise sequences for an autonomous vehicle navigating a four-way intersection based on the current relative position and velocity of an intruder. Using the generative adversarial architecture, the 1000-step model is distilled into a single-step denoising diffusion model which demonstrates fast inference speed while maintaining similar sampling quality. We demonstrate one possible application of the single-step model in building a robust planner for the autonomous vehicle. The planner uses the single-step model to efficiently sample potential failure cases based on the currently measured traffic state to inform its decision-making. Through simulation experiments, the robust planner demonstrates significantly lower failure rate and delay rate compared with the baseline Intelligent Driver Model controller.
\end{abstract}

\section{Introduction}
\label{intro}
Traffic intersections have remained a significant safety bottleneck. In 2022, almost 30\% of traffic fatalities involve an intersection, and in 2023, over 50\% of crashes resulting in fatalities or injury occur at or near intersections \cite{2022fatalities,2023crashes}. Due to the need to constantly monitor approaching vehicles and predict their intentions accurately, it is extremely challenging to safely navigate an intersection. In recent years, autonomous vehicle (AV) technology has made substantial progress in terms of safety and reliability. However, recent statistics still reveal that intersections are responsible for 55\% of driverless car accidents, and there has been growing interest in developing robust autonomous driving systems that effectively navigate these high-risk areas \cite{jobera2024driverless}.

This study develops a robust planning framework for an autonomous vehicle navigating a four-way intersection in the presence of an intruder vehicle. Our planning framework is ``robust'' in the sense that it is aware of the potential failures given the current state. Recent AV robust planning methods seek to mitigate failure-causing vulnerabilities including unseen environments as well as observation and tracking inaccuracies \cite{yu2022formally, zeng2019novel, unal2023towards}. To the best of our knowledge, few of the frameworks proposed so far comprehensively account for the overall distribution of potential failures given the current relative positions and velocities of intruder vehicles. 

Inspired by recent safety validation frameworks, we use a denoising diffusion model to represent the broader distribution of potential failures of the autonomous vehicle. We train diffusion models to sample collision-causing observation error sequences based on the current state \cite{jun, delecki2024diffusion}. We perform knowledge distillation for the diffusion models to accelerate inference, enabling real-time generation of failure samples.

The robust planning framework uses the distilled diffusion model to efficiently draw potential failure samples based on the real-time traffic situation at each decision step and considers a variety of potential failures during decision-making. In particular, the planner computes a trajectory to its destination that avoids the potential collisions. Our code is available on GitHub.\footnote{https://github.com/sisl/AV-Robust-Planning-Diffusion-Failure-Sampler} Our main contributions include:
\begin{itemize}
    \item We build a \textbf{real-time failure sampler} for an autonomous vehicle at an intersection, which rapidly generates potential failure trajectories based on the current relative position and velocity of an intruder vehicle.
    \item We use the real-time failure sampler to construct a \textbf{robust planner} which accounts for the \textbf{overall distribution of potential failures} for reliable decision-making.
\end{itemize}

\begin{figure}[t]
    \centering
    \includegraphics[width=\linewidth]{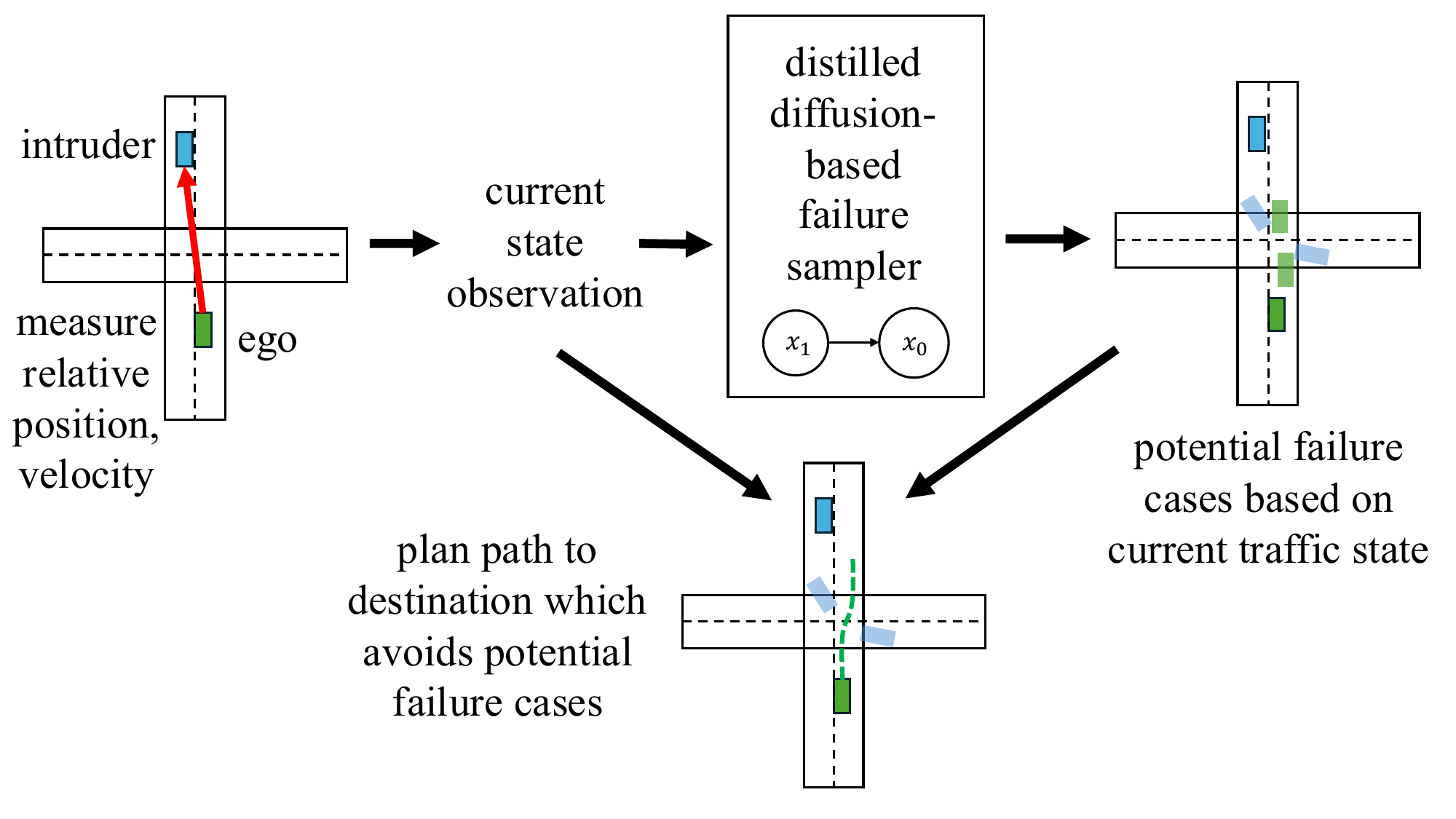}
    \caption{Our robust planner uses a distilled denoising diffusion model to sample potential failure cases based on the currently measured traffic state. It plans a path to the destination which avoids the potential failure cases.}

\end{figure}

\Cref{Related Work} provides an overview of current research in AV robust planning. \Cref{Methods} discusses the training of the \newgeometry{top=0.7in, bottom=0.7in, left=0.7in, right=0.7in} failure sampler and the design of the robust planner. \Cref{Experiments} presents the experimental results for the real-time failure sampler and the robust planner.

\section{Related Work}\label{Related Work}

Recent studies have combined techniques from both safety validation and trajectory planning to address a variety of uncertainties in autonomous driving. To adapt the autonomous vehicle to complex environments, Jaafra \textit{et al.} \cite{jaafra2019robust} trained a deep reinforcement learning (RL) agent with an actor-critic architecture and multi-step returns, ensuring robustness against realistic urban driving conditions. Sun \textit{et al.} \cite{sun2022robust} developed a Cartesian-to-Frenet projector that enables Frenet-based lane keeping on high-curvature roads. 

To better address unseen traffic scenarios during real-world operations, monitoring controllers that prioritize safety under special circumstances have been designed. Zeng \textit{et al}. \cite{zeng2019novel} developed a controller that defaults to a monitoring trajectory that ensures collision avoidance when a lane changing trajectory cannot be computed. Yang \textit{et al.} \cite{yang2023towards} proposed a hybrid controller in which a rule-based policy takes over when the RL policy encounters a scenario not seen during training. To enhance robustness against tracking deviations, Kalaria \textit{et al.} \cite{kalaria2022delay} designed a framework using robust tube MPC and and adaptive Kalman Filter to deal with actuator delays, Yu \textit{et al}. \cite{yu2022formally} proposed a feedback control technique that bounds the tracking error of the actual trajectory to a geometric tube, and Zhang \textit{et al}. \cite{zhang2021robust} developed a gain-scheduling approach for robust lateral tracking.

There has been growing interest in improving the robustness of AV planners by considering observation errors, especially those that are likely to cause failures. He \textit{et al}. proposed a framework in which a Bayesian Optimization-based adversarial attacker introduces worst-case observation perturbations in an attempt to produce a collision, which helps an RL agent learn an optimal policy that is robust against risky observation errors \cite{he2023towards, he2024trustworthy}. Unal \textit{et al}. proposed a genetic algorithm that introduces noise to the training data for object classifiers onboard an AV, making the classifiers more robust against data perturbations \cite{unal2023towards}. 

Nevertheless, these approaches account for relatively small sets of worst-case perturbations most likely to lead to failures. Few studies to date have comprehensively modeled the full distribution of potentially failure-causing perturbations for the purpose of robust planning. We believe the latter approach is advantageous because it informs the autonomous vehicle of a wider range of potential failures based on the current traffic state, enabling more robust decision-making in real time. Recent safety validation frameworks have demonstrated the ability of diffusion models to model the overall distribution of failure-causing observation perturbations for many cyber-physical systems \cite{delecki2024diffusion, jun}. In our study, we decide to develop a robust planning framework for an autonomous vehicle in a traffic intersection environment that uses a distilled diffusion model to obtain knowledge about a variety of likely failures based on the current traffic state.

\section{Methods} \label{Methods}
\subsection{Conditional Diffusion-Based Failure Sampling}
\label{setup}
First, we build a generative model which synthesizes collision-causing observation disturbance sequences for the autonomous vehicle. Consider an autonomous vehicle navigating an intersection\footnote{https://github.com/Farama-Foundation/HighwayEnv} with four road branches: south (bottom), north (top), west (left), and east (right).  Each has two lanes for the two directions of travel. The ego vehicle must travel from the south branch to one of the other branches while avoiding an intruder. The intruder vehicle might emerge from any branch and might turn left, go straight, or turn right. The initial positions and velocities of the vehicles are randomized. For the purpose of failure sampling, we assume both vehicles are controlled by the Intelligent Driver Model (IDM) \cite{treiber2000congested} within lanes. The ego vehicle follows a fixed policy, while the intruder exhibits randomized behavior with the delta-exponent of its IDM policy randomly uniformly sampled between 3.5 and 4.5.

\begin{figure}[t]
  \includegraphics[width=\columnwidth]{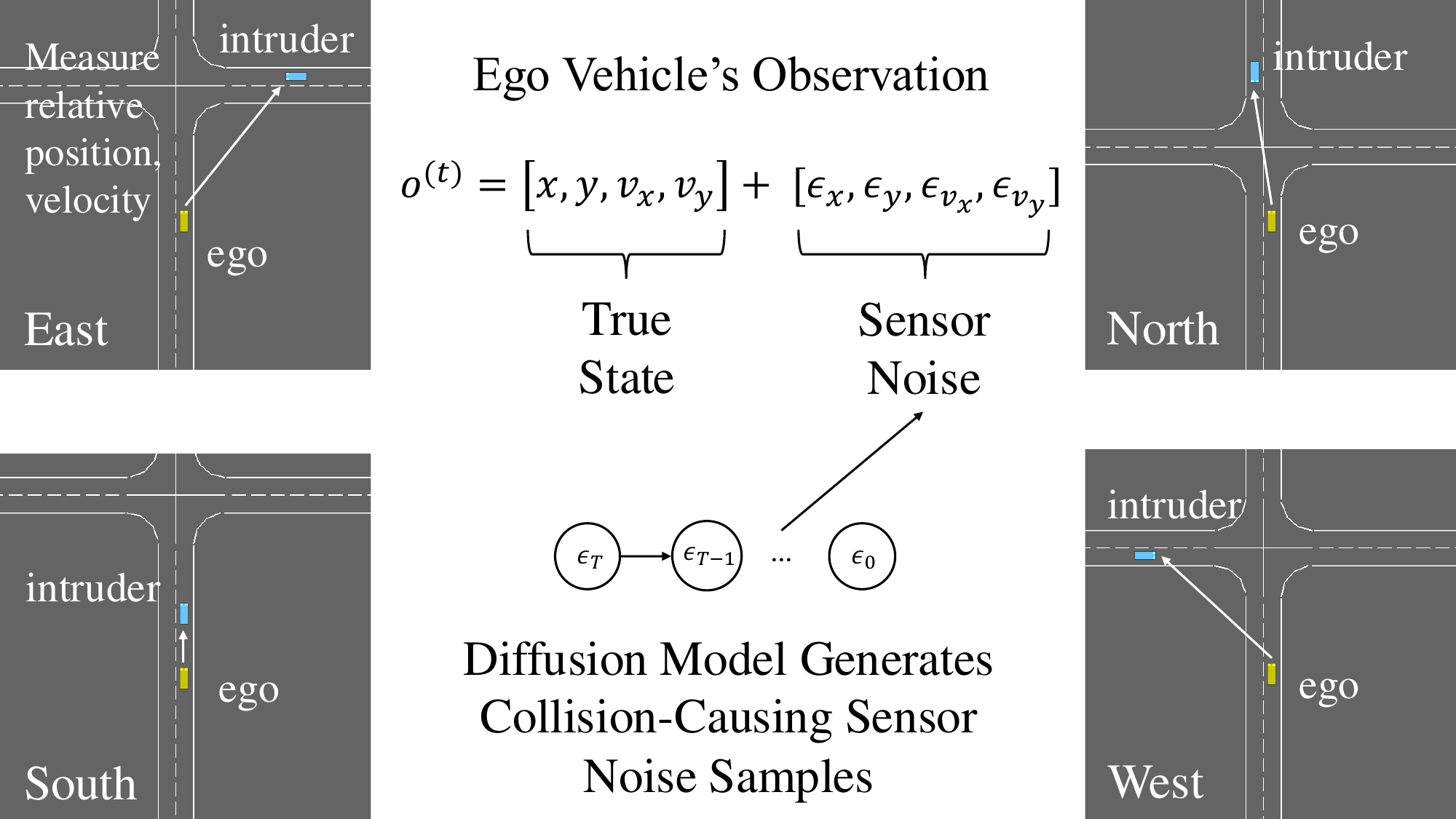}
  \caption{Illustration of our problem setup. A self-driving ego vehicle attempts to navigate an intersection in the presence of an intruder vehicle that may approach the intersection from the east, west, south, and north road branch.}
  \label{problem viz}
\end{figure}

The ego vehicle has no information on the intruder's destination road branch. Its sensor measures the intruder's position and velocity $x_{\text{intr}}, y_{\text{intr}}, v_{x,\text{intr}}, v_{y, \text{intr}}$ through a sensor with a Gaussian prior noise model, $\epsilon^{(t)} = [\epsilon_x, \epsilon_y, \epsilon_{v_x}, \epsilon_{v_y}] \sim \mathcal{N}(\mathbf{0}, \gamma \mathbf{I})$, where $\gamma$ is the noise scale. Based on each observation, the IDM controller computes the acceleration to perform over the next timestep along its planned path of motion, with no lateral motion. It performs a fixed horizon of 23 observations and actions starting from the initial timestep to complete a \textit{simulation}. The \textit{robustness} $\rho$ of a simulation refers to the minimum separating distance between the vehicles throughout the simulation. A $\rho$ of 0 corresponds to a collision or \textit{failure}. A \textit{scenario} refers to the set of possible situations with the intruder emerging from a specific road branch. There are four scenarios in total, as shown in \cref{problem viz}. In each scenario, given the initial state of the system $\textbf{s}_{0}$, and a particular sequence of observation errors for the ego $\bm{\epsilon} = [\bm{\epsilon^{(t=1)}}, \dots, \bm{\epsilon^{(t=23)}}]$, a set of different simulations may occur. The robustness of an instance of a scenario is therefore a random variable conditioned on the initial state and the observation error, $\rho(\textbf{s}_{0}, \bm{\epsilon})$. The task of failure sampling can thus be formalized as follows: for each scenario $\{\text{east}, \text{west}, \text{south}, \text{north}\}$, given the initial positions and velocities of the ego and intruder, $\textbf{s}_{0}$, we want to sample collision-causing observation error sequences $\bm{\epsilon}$ from:\begin{equation}\label{targetdistribution}
    p(\bm{\epsilon} \mid \rho=0, \mathbf{s}_{0}) \propto p(\rho(\mathbf{s}_{0}, \bm{\epsilon})=0)p(\bm{\epsilon})
\end{equation}where $p(\bm{\epsilon})$ has a Gaussian prior $\mathcal{N}(\bm{\epsilon}; \mathbf{0}, \gamma \mathbf{I})$. Note that this implies $\bm{\epsilon}$ is independent of the initial state $\mathbf{s}_0$ 

We adopt the diffusion-based failure sampling framework developed recently \cite{delecki2024diffusion, jun}. The diffusion model transforms a unit Gaussian  $\bm{\epsilon}_K$ to a sample $\bm{\epsilon_0}$ from the distribution given by \cref{targetdistribution} over $K$ denoising steps parameterized as $\mathbf{\mu}_{\psi}$. It samples from the distribution with density $p_{\psi}(\bm{\epsilon}_{0:K} \mid \rho_{\text{threshold}}, \mathbf{s}_0)$, which is given by 
\begin{equation} \label{diffusionmodel}
    p(\bm{\epsilon}_K)\prod_{k=1}^{K} \mathcal{N}(\bm{\epsilon}_{k-1} \mid \mathbf{\mu}_{\psi}(\bm{\epsilon}_{k} \mid k, \rho_{\text{threshold}}, \mathbf{s}_0), \beta_k \mathbf{I})
\end{equation}
where $p(\bm{\epsilon}_K) = \mathcal{N}(\bm{\epsilon}_K \mid \mathbf{0}, \gamma \mathbf{I})$ because the prior noise model is Gaussian. The variance schedule $\beta_1, \dots, \beta_k$ is a fixed increasing sequence based on the cosine schedule. The robustness threshold $\rho_{\text{threshold}}$ specifies the desired robustness of simulations with the ego experiencing the generated observation error. For simplicity, we denote the process of generating failure sample $\bm{\epsilon}_0$ as  $\bm{\epsilon}_0 \sim p_{\psi}(\bm{\epsilon}_K, \rho_{\text{threshold}}, \mathbf{s}_0)$.  During inference, we set $\rho_{\text{threshold}}=0$ to generate observation error samples that lead to collisions (i.e., $\rho=0$). We initially tried to train a single model across all four scenarios. However, the training data was quickly dominated by failure samples from the North scenario, where failures are significantly more frequent. This imbalance caused the model to overlook failure modes involving intruder vehicles originating from other branches. As a result, we opted to train a separate diffusion model for each scenario. The training algorithm is explained in more detail in \cite{jun, delecki2024diffusion}.

\subsection{GAN-Based Knowledge Distillation}
\label{knowledge_distillation}
The diffusion model generates failure samples through a large number of $K=1000$ denoising steps. This autoregressive denoising process leads to long inference time. To enable applications in real-time robust planning, we apply knowledge distillation to the diffusion model to accelerate inference. We want to distill the knowledge of the teacher model $p_{\psi}(\bm{\epsilon}_{0:K} \mid \rho_{\text{threshold}}, \mathbf{s}_0)$ to a one-step student model. The student model has variance schedule $\hat{\beta}$ computed using the cosine schedule, and it uses a single denoising step to generate failure sample $\bm{\epsilon}_0$, with density $p_{\theta}(\bm{\epsilon}_{0} \mid \rho_{\text{threshold}}, \mathbf{s}_0)$ which is given by: \begin{equation} p(\bm{\epsilon}_{1})\mathcal{N}(\bm{\epsilon}_{0} \mid \mathbf{\mu}_{\theta}(\bm{\epsilon}_{1} \mid \rho_{\text{threshold}}, \mathbf{s}_0), \hat{\beta} \mathbf{I})
\end{equation} where $p(\bm{\epsilon}_{1}) = \mathcal{N}(\bm{\epsilon}_1 \mid \mathbf{0}, \gamma \mathbf{I})$. For simplicity, we denote the student model's generation process as $\hat{\bm{\epsilon}} = \bm{\epsilon}_0 \sim p_{\theta}(\bm{\epsilon}_1, \rho_{\text{threshold}}, \mathbf{s}_0)$. We adopt the distillation pipeline proposed by recent studies on diffusion distillation  \cite{sauer2025adversarial}. 

We first draw a large sample dataset from the teacher model. For each scenario $\{\text{east, west, south, north}\}$ we initialize $N$ simulations with initial states $\mathbf{s}_0^{(1)}, \dots, \mathbf{s}_0^{(N)}$ as well as Gaussian samples $\bm{\epsilon}_K^{(1)}, \dots, \bm{\epsilon}_K^{(N)}$, and then use the teacher model to draw collision-causing observation error samples for each initial state: $\bm{\epsilon}^{(j)} \sim p_{\psi}(\bm{\epsilon}_K^{(j)}, 0, \mathbf{s}_0^{(j)}), j=1, \dots, N$. We then run the simulation environments while subjecting the ego vehicle to the generated observation errors, record the robustness values of the simulations, and then construct a dataset $\mathcal{D} = \{(\bm{\epsilon}^{(j)}, \rho^{(j)}, \mathbf{s}_0^{(j)}), j=1, \dots, N\}$.

The student model is first pretrained with the same algorithm used to train the teacher diffusion models until convergence \cite{jun}. Subsequently, supervised pretraining is performed with the teacher dataset $\mathcal{D}$. Let $(\bm{\epsilon}, \rho, \mathbf{s}_0)$ be a sample from $\mathcal{D}$. We diffuse $\bm{\epsilon}$ using the student model's forward process to produce noised data point $\Tilde{\bm{\epsilon}} \sim \mathcal{N}(\sqrt{1-\hat{\beta}}\bm{\epsilon}, \hat{\beta} \mathbf{I})$. The student model is then asked to reconstruct the original teacher sample $\bm{\epsilon}$ from the noised data $\tilde{\bm{\epsilon}}$, and it outputs $\hat{\bm{\epsilon}} \sim p_{\theta}(\tilde{\bm{\epsilon}}, \rho=0, \mathbf{s}_0)$. We then compute the Mean-Squared Error Loss (MSE) between $\bm{\epsilon}$ and $\hat{\bm{\epsilon}}$:
\begin{equation}\label{supervised_pretraining}
    \mathcal{L}_{supervised} = MSE(\bm{\epsilon}, \hat{\bm{\epsilon}})
\end{equation}

The main distillation step uses a Generative Adversarial Network (GAN) architecture. Again taking a teacher sample $(\bm{\epsilon}, \rho, \mathbf{s}_0) \in \mathcal{D}$, we have the student model diffuse and then reconstruct the sample $\bm{\epsilon}$ similar to the previous step, and it outputs $\hat{\bm{\epsilon}}$. The teacher sample $\bm{\epsilon}$ and the student reconstruction $\Hat{\bm{\epsilon}}$ are then passed to a discriminator model $M_{\phi}$, which aims to distinguish between the two. In this study, we choose a ResNet architecture \cite{he2016deep} for the discriminator, similar to each downsampling and upsampling layer of the U-Net used by the diffusion model. The generator is trained to fool the discriminator into classifying its reconstruction as the original sample (i.e. predicting a score close to 1). The \textit{adversarial loss} for the student model is therefore:
\begin{equation}\label{adversarial_loss}
    \mathcal{L}^G_{adv}(\hat{\bm{\epsilon}}, \phi) = -\log M_{\phi}(\hat{\bm{\epsilon}})
\end{equation} whereas the adversarial loss for the discriminator is:
\begin{equation}\label{discriminator_adv_loss}
    \mathcal{L}^M_{adv}(\bm{\epsilon}, \hat{\bm{\epsilon}}, \phi) = -\log (1-M_{\phi}(\hat{\bm{\epsilon}})) -\log{M_{\phi}(\bm{\epsilon})}
\end{equation}

In order to train the student model to imitate the denoising behavior of the teacher model, we combine the adversarial loss with an additional \textit{distillation loss}. To compute this loss, we diffuse the student's reconstructed sample $\hat{\bm{\epsilon}}$ using the teacher model's forward process to $\hat{\bm{\epsilon}}_k \sim \mathcal{N}(\sqrt{1 - \beta_k}\hat{\bm{\epsilon}}, \beta_k \mathbf{I})$ where $k$ is sampled uniformly from $\{1, \dots, K\}$. We then use the teacher model to reconstruct the original sample based on $\hat{\bm{\epsilon}}_k$, obtaining the output $\hat{\bm{\epsilon}}_{\psi} \sim p_{\psi}(\hat{\bm{\epsilon}}_k, 0, \mathbf{s}_0)$. The distillation loss measures the mismatch between the student output $\hat{\bm{\epsilon}}$ and the teacher output $\hat{\bm{\epsilon}}_{\psi}$: \begin{equation}\label{distillation_loss}
    \mathcal{L}_{distill}(\hat{\bm{\epsilon}}, \hat{\bm{\epsilon}_{\psi}}) = \| \hat{\bm{\epsilon}} - \hat{\bm{\epsilon}_{\psi}} \|_2^2
\end{equation} and thus the overall objective for the generator is \begin{equation}\label{loss}
    \mathcal{L} = \mathcal{L}^G_{adv}(\hat{\bm{\epsilon}}, \phi) + \lambda \mathcal{L}_{distill}(\hat{\bm{\epsilon}}, \hat{\bm{\epsilon}_{\psi}})
\end{equation}

In each training iteration, we draw a batch of samples from $\mathcal{D}$, update the discriminator based on the loss given in \cref{discriminator_adv_loss}, and then update the student model based on the loss given in \cref{loss} in an alternating fashion.

\subsection{Robust Planning with Failure Samples} \label{Planning Method}
For the purpose of failure sampling, we assume that both vehicles are controlled by the IDM. With access to failure samples generated by the student (distillation) model from \cref{knowledge_distillation}, we aim to construct a more robust controller for the ego vehicle while still assuming an IDM-controlled intruder. We use an iterative optimization-based planning method: at each decision step, the distillation model trained for the current scenario is used to generate collision-causing sensor noise samples. Traffic simulations are performed to determine the state trajectories the vehicles would undergo if they were controlled by the IDM while experiencing the generated sensor noise sequences. We call these high-risk state trajectories our \textit{failure samples}. It is desirable for the ego vehicle to stay clear of any potential high-risk trajectory of the intruder vehicle. To achieve this, we use linear programming (LP) to compute the action sequence that maximizes the worst-case road distance between the ego vehicle and a potential intruder position across all future timesteps and failure samples. Appendix \ref{Planner Details} discusses the technical details of our optimization-based planning framework. Representing the potentially risky future behavior of the intruder with a fixed number of failure samples improves scalability because it does not require simulating the planning processes of and interactions between the intruders online. In a potential extension of our framework to scenarios with multiple intruders, the number of constraints in our LP scales linearly with the number of intruders.

The ego vehicle performs this optimization procedure iteratively at each timestep and carries out the next optimal action. However, as explained in \cref{Experiments}, the diffusion and distillation models are trained for situations where the vehicles are relatively far from the intersection. When the vehicles travel too close to the intersection, we are unable to generate new failure samples. In this case, we reuse the most recent batch of failure samples generated. Each of these samples, besides the sensor observation, corresponds to a belief about a possible state trajectory of the intruder. We use Kalman Filters to represent these beliefs. We sample possible intruder states from each filter and select ``plausible'' intruder states. An intruder state is considered plausible if its positional component is within $\eta$ of the current observed intruder position. The IDM policy is used to compute an action for each plausible intruder state, and the most conservative (lowest-acceleration) action is performed. We refer the reader to appendix \ref{Planner Details} for a more formal description of the planning algorithm.

\subsection{Evaluation Metrics and Baselines}
To determine the effectiveness of the knowledge distillation framework, we compare the sampling quality of the failure samplers (i.e. distillation models) against that of their teacher diffusion models. Sampling quality is defined as the extent to which the simulations generated by the failure sampler match the true distribution of potential failures in the scenario of interest. By ``simulations generated by the failure sampler," we refer to the simulations (i.e. vehicle state trajectories) that occur as a result of the observation error sequences generated by the failure sampler. Based on an established framework for evaluating generative models \cite{naeem2020reliable}, we propose the following evaluation metrics:

\begin{itemize}
    \item \textbf{failure rate ($\uparrow$)}: Fraction of actual collisions among the simulations generated. A high failure rate indicates less waste of computing in synthesizing non-collision samples, i.e. a sample-efficient generation of failures. This metric is inclusively bounded between 0 and 1. As opposed to the robust planner, it is desirable for the failure sampler to achieve a high failure rate.
    \item \textbf{density ($\uparrow$)}: This metric measures how well the simulations generated match the prevalent failure modes in the true failure distribution. A higher density indicates that the model synthesizes more realistic failures. This metric is unbounded and can exceed 1.
    \item \textbf{coverage ($\uparrow$)}: This metric measures how well the simulations generated capture the full range of variability in the true failure distribution. A high coverage means the model can generate more diverse failure samples. This metric is inclusively bounded between 0 and 1.
\end{itemize}

To evaluate the robust planner, we assess its ability to transport the ego vehicle to its destination without colliding with the intruder and within the horizon of 23 action steps, as compared with the IDM controller used for the purpose of failure sampling. Therefore, the following evaluation metrics are used for the robust planner:

\begin{itemize}
    \item \textbf{failure rate ($\downarrow$)}: The probability that the robust planner causes the ego vehicle to collide with the IDM-controlled intruder vehicle. As opposed to the failure sampler, the robust planner aims for a low failure rate.
    \item \textbf{delay rate ($\downarrow$)}: The probability that the ego vehicle fails to reach its destination within 23 action steps when controlled by the robust planner.
\end{itemize}

\section{Experiments and results} \label{Experiments}
\subsection{Environment Setup and Monte Carlo Simulations}
In our simulation environment, the ego vehicle's initial distance to the intersection is sampled from $\mathcal{U}[0.35, 0.65]$ and its initial speed from $\mathcal{U}[0.35, 0.5]$ (in units per second), where each lane is 0.04 units wide. The intruder's initial distance to the intersection is sampled from $\mathcal{U}[0.25, 0.45]$ and its speed from $\mathcal{U}[0.35, 0.45]$. The destination road branches of the vehicles and the behavior of the intruder are randomized based on HighwayEnv's provided feature. We set the prior sensor noise scale $\gamma = 1/0.15$. We first construct an IDM-controlled ego vehicle and performed roughly $8 \times 10^8$ Monte Carlo simulations over a period of 3 weeks. We obtained 1028, 1892, 1083, and 1017 Monte Carlo failure samples for the South, North, West, and East scenarios, respectively. These samples approximately represent the true failure distribution of the system in the four scenarios and are thus used to evaluate the diffusion-based failure samplers.

\subsection{Diffusion-Based Failure Sampler: Training and Distillation}
Next, we trained and distilled diffusion models to generate failure samples for the intersection system, assuming both vehicles are IDM-controlled. Our diffusion model is based on a U-Net with four downsampling and upsampling layers, which we implemented using PyTorch. A separate model with $K=1000$ denoising steps was trained for each scenario using the AdamW optimizer \cite{delecki2024diffusion}. We then drew over 1000 failure samples from the model for each scenario and compared the diffusion samples against the Monte Carlo failure samples to evaluate the sample fidelity of the models.

The 1000-step failure samplers for the four scenarios (i.e. teacher models) were then distilled to 1-step distillation (student) models. First, we drew a dataset containing 200,000 samples from each teacher model. The student models were first pretrained using the cross-entropy method \cite{jun,delecki2024diffusion} and then subjected to supervised pretraining with the teacher datasets, with a batch size of 256, a learning rate of $10^{-3}$, an L2-regularization strength of $10^{-5}$, and 50,000 update steps. In the GAN-based knowledge distillation stage, the discriminator is constructed using the same ResNet architecture as that used in the U-Net for the diffusion model, with an additional linear prediction layer. We chose to weigh the adversarial and distillation losses equally, setting $\lambda = 1$ as in \cref{loss}. The learning rate for both the generator and discriminator was set to $3 \times 10^{-4}$. The regularization strength was set to 0.6 for the generator and 0.1 for the discriminator, and the batch size was set to 2048. The overall distillation process for each scenario took roughly 36 hours on a NVIDIA GeForce RTX 4090 GPU. The sampling quality the student models compared with their teachers are shown in \Cref{distillation_results}. We recorded the inference times of the diffusion model and the distillation model for 1000 samples on a GeForce RTX 4090 GPU over 10 repeated experiments, and the results are shown in \Cref{inference_speeds}.

As shown in \cref{distillation_results}, in the East, West, and South scenarios, the density and coverage performances of the distillation models are close to those of their teacher models, showcasing that the sampling distribution of the distillation models is similarly close the true failure distribution as that of the teacher models. In the North scenario, the distillation model achieves a significantly higher failure rate than its teacher and yet falls short in terms of density and coverage. This suggests that the distillation model may be focused on generating a smaller set of most likely failures while sacrificing the diversity of failures of the teacher model. Finally, as seen in \cref{inference_speeds}, the average inference speed of the distillation model is substantially faster than that of the diffusion model.

\begin{table}[ht]
    \centering    \caption{Performance metrics for the diffusion model and the distillation model in all four scenarios.}
    \sisetup{detect-weight=true, detect-family=true}
    \begin{adjustbox}{max width=\columnwidth}
    \begin{tabular}{@{}llS[table-format=1.4]
    S[table-format=1.4]
    S[table-format=1.4]@{}}
        \toprule
        \textbf{Scenario} & \textbf{Model} & \textbf{Failure Rate $\uparrow$} & \textbf{Density $\uparrow$} & \textbf{Coverage $\uparrow$} \\
        \midrule
        \multirow{2}{*}{\parbox{1cm}{East}}   & Diffusion model & 0.0160 & \num{0.8924} &  \num{0.7837} \\
               & Distillation model            &  0.0164 & \num{0.9390} & \num{0.5841} \\
        \midrule
        \multirow{2}{*}{\parbox{1cm}{West}}    & Diffusion model & \num{0.0094} &  \num{0.8676} &  \num{0.4314} \\
               & Distillation model            & \num{0.0140} & \num{0.7217} & \num{0.3139} \\
        \midrule
        \multirow{2}{*}{\parbox{1cm}{North}}   & Diffusion model & \num{0.0048} & \num{0.9558} & \num{0.8967} \\
               & Distillation Model             & \num{0.0232} & \num{0.6436} & \num{0.2045} \\
        \midrule
        \multirow{2}{*}{\parbox{1cm}{South}}  & Diffusion model & \num{0.2134} & \num{0.9002} & \num{0.8045} \\
               & Distillation Model             & \num{0.2670} & 0.8327 & 0.7442 \\
        \bottomrule
    \end{tabular}
    \end{adjustbox}
    \label{distillation_results}
\end{table}

\begin{table}[ht]
    \centering    \caption{Means and standard deviations of the inference speeds of the 1000-step diffusion model and the 1-step distillation model}
    \sisetup{detect-weight=true, detect-family=true}
    \begin{adjustbox}{max width=\columnwidth}
    \begin{tabular}{@{}lr@{}}
        \toprule
        \textbf{Model} & \textbf{Inference Time (seconds per 1000 samples) $\downarrow$}  \\
        \midrule
        Diffusion model & \num{24.98} $\pm$ \num{0.019} \\
        Distillation model & \bfseries \num{0.280} $\pm$ \num{0.009} \\
        \bottomrule
    \end{tabular}
    \end{adjustbox}
    \label{inference_speeds}
\end{table}

\subsection{Robust Planning with Failure Sampler} \label{planner_experiments}
The distillation models are able to generate potential failure samples in real time for situations wherein both vehicles are IDM-controlled. These models are then used to build a more robust planner for the ego vehicle while still assuming an IDM-controlled intruder vehicle. At each timestep of the planning phase, the distillation model trained for the current scenario is used to generate $N' = 200$ samples, and the $N=40$ lowest-robustness samples are selected for use as failure samples by the planner. After 23 actions, the ego vehicle is considered to have reached its destination if it is located on the target road branch, with a distance of at least 0.1 from the intersection. For the trajectory optimizer, we set the maximum forward velocity in each cardinal direction to 0.5 and the maximum retrograde velocity to $10^{-3}$. The maximum forward acceleration is set to 0.1 in the x-direction (East/West) and 1.0 in the y-direction, while the maximum retrograde acceleration is set to 0.2 in all directions. The maximum mean velocity is set to 0.4 in the x-direction and 0.3 in the y-direction. During the policy-based phase, we consider a failure sample ``plausible" and pass it to the IDM policy if, at the current timestep, the intruder location in the trajectory is within $\eta = 0.08$ of the currently observed intruder position.

\begin{table}[H]
    \centering    \caption{Performance metrics for the robust planner and the Intelligent Driver Model (IDM) in all four scenarios.}
    \sisetup{detect-weight=true, detect-family=true}
    \begin{adjustbox}{max width=\columnwidth}
    \begin{tabular}{@{}lrrr
    S[table-format=1.4]@{}}
        \toprule
        \textbf{Planner} & \textbf{Simulations Performed} & \textbf{Failure Rate $\downarrow$} & \textbf{Delay Rate $\downarrow$} \\
        \midrule
        Robust  Planner & \num{130000} & \bfseries \num{0.846e-4} & \bfseries \num{0.856} \\
         IDM            & \num{130000} &  \num{2.231e-4} & \num{0.867}\\
        
        \bottomrule
    \end{tabular}
    \end{adjustbox}
    \label{planner_results}
\end{table}

We evaluate the performance improvement of the robust planner compared with the baseline IDM controller used in the context of failure sampling. Simulations are performed for two controller configurations: (1) IDM-controlled ego vehicle with the same configuration as that used for failure sampling, and (2) robust planner-controlled ego vehicle. In both cases, the intruder is IDM-controlled with randomized behavior, as in previous experiments. The delay and collision rates are presented in \Cref{planner_results}. We applied a one-tailed two-proportion $z$-test to assess whether the robust planner exhibits a statistically significantly lower failure rate and delay rate. We found that the robust planner demonstrates a lower failure rate than the IDM, with a $p$-value of $0.002$. The robust planner also demonstrates a lower delay rate, with a $p$-value of $<0.001$. For both performance metrics, the $p$-value is well below the conventional significance threshold of $0.05$, and thus we conclude that the robust planner demonstrates a significantly lower failure rate and delay rate than the IDM. This means that, compared with the IDM, the robust planner is significantly less likely to cause collisions and more likely to reach the destination within 23 action steps.

\section{Conclusion}
This study applies deep generative models to robust planning in autonomous driving. Inspired by recent diffusion-based safety validation frameworks, we train diffusion models to generate collision-causing observation error samples for an autonomous vehicle in a traffic intersection based on the current traffic state. We demonstrate that the diffusion models can be distilled into student models with much higher sampling speeds while maintaining comparable sampling quality. We propose a robust planning framework for an autonomous vehicle in the same intersection context that uses the distillation models to generate potential failure samples in real time. These failure samples are used to inform the decision-making process. Based on our experimental results, we conclude that the robust planner achieves a significantly lower failure rate and delay rate compared with the baseline IDM controller.

In the future, we plan to generalize our robust planning framework to account for traffic situations with multiple intruder vehicles. Since our framework does not rely upon any prior assumptions about the road geometry of interest, we may also experiment on other types of safety-critical traffic environments, such as roundabouts and highway merging. Since generative modeling is a rapidly advancing field, we believe that future generative architectures may be able to model the failure distribution of an autonomous vehicle more effectively in a wider variety of traffic scenarios, further enhancing the safety and reliability of autonomous driving.
\label{Conclusions}

\vspace{-0.3cm}

\bibliographystyle{ieeetr}
\bibliography{references}

\appendices

\section{Technical Design of the Robust Planner} \label{Planner Details}
\subsubsection{Planning Phase}
Given the initial state $\mathbf{s}_0$ and a sequence of control inputs $a_1, \dots, a_{23}$, the system will undergo the state trajectory $\mathbf{S}=[\mathbf{s}_0, \mathbf{s}_1, \dots, \mathbf{s}_{23}]$.  Suppose we are at timestep $t$. The sensor obtains a noisy measurement of the intruder state $\mathbf{o}_t = [x_{\text{intr}}, y_{\text{intr}}, v_{x, \text{intr}}, v_{y, intr}]$. We construct $N'$ simulation environments with initial state $\mathbf{o}_t$ and a horizon of $23-t$ action steps, wherein both vehicles are assumed IDM-controlled. The distillation model trained for the current scenario is used to generate collision-causing sensor noise samples $\hat{\bm{\epsilon}}^{(1)}, \ldots, \hat{\bm{\epsilon}}^{(N')}$. The environments then perform simulations with these sensor noise samples, recording the resulting state trajectories $\mathbf{F}^{(1)}, \ldots, \mathbf{F}^{(N')}$ and robustness values $\rho^{(1)}, \dots, \rho^{(N')}$, where $\mathbf{F} = [\mathbf{o}_t, \mathbf{f}_{t+1}, \mathbf{f}_{t+2}, \dots, \mathbf{f}_{23}]$, $\mathbf{f}=[x_{\text{intr}}, y_{\text{intr}}, v_{x, \text{intr}}, v_{y, \text{intr}}]$. The $N$ state trajectories with the lowest robustness values are used as failure samples. At any future timestep $t' \in \{t+1, t+2, \dots, 23\}$, across the failure samples, the intruder may be located at different positions $(\mathbf{f}_{t'}^{(j)})_{[1:2]}, j=1,\dots,N$. It is desirable for the ego vehicle to stay clear of these positions as much as possible. To achieve this, we compute the action sequence $a_{t+1}, \dots, a_{23}$ that maximizes the minimum road distance between the ego vehicle and a potential intruder vehicle position across all future timesteps and failure samples. This optimization is performed by solving a linear program using gurobipy:\footnote{https://www.gurobi.com}\\

\text{Maximize} 

\begin{align}
    \min_{t', j} |x_{ego, t'} - (\mathbf{f}_{t'}^{(j)})_1| + |y_{ego, t'} - (\mathbf{f}_{t'}^{(j)})_2| \notag\\
    \text{for } t' \in \{t+1, \dots, 23\}, 
    j\in\{1, \dots, N\} \notag
\end{align}

\text{subject to}
\begin{align*}
    x_{ego, t}, y_{ego, t} &= \text{current ego position}  \\
\end{align*}
\begin{align*}
    v_{x, ego, t}, v_{y, ego, t} &= \text{current ego velocity} \\
    (x_{ego, 23}, y_{ego, 23}) &\in \mathcal{Q} \notag\\
    \text{and for all future timesteps } t' &\in \{t+1, \dots, 23\}:\\
    (x_{ego, t'}, y_{ego, t'}) &\in \mathcal{L} \\
    x_{ego, t'+1} &= x_{ego, t'} + v_{x, ego, t'} + \frac{1}{2} a_{t', x}\\
    v_{x, ego, t'+1} &= v_{x, ego, t'} + a_{t', x} \\
     y_{ego, t'+1} &= y_{ego, t'} + v_{y, ego, t'} + \frac{1}{2} a_{t', y}\\
    v_{y, ego, t'+1} &= v_{y, ego, t'} + a_{t', y} \\
    v_{x, ego, t'} &\in [v_{x, min}, v_{x, max}]\\
    v_{y, ego, t'} &\in [v_{y, min}, v_{y, max}] \\
    \mathop{\mathbb{E}}_{t'}[|v_{x, ego, t'}|] &\leq \bar{v}_{x, max}\\
    \mathop{\mathbb{E}}_{t'}[|v_{y, ego, t'}|] &\leq \bar{v}_{y, max}\\
    a_{t', x} &\in [a_{x, min}, a_{x, max}]\\
    a_{t', y} &\in [a_{y, min}, a_{y, max}]\\
\end{align*} 

$\mathcal{Q}$ denotes the destination region with linear boundaries, as specified in \cref{planner_experiments}. $\mathcal{L}$ denotes the lane(s) within which the ego vehicle is allowed to travel throughout the journey, with linear boundaries. The LP formulation uses a linear simplification of the point-mass model with forward Euler integration, ensuring consistency with the dynamics model implemented in HighwayEnv (our simulation environment). We assume the ego vehicle accurately observes its own state.

\subsubsection{Policy-Based Phase}
As explained in \cref{Planning Method}, when the vehicles travel too close to the intersection, we are unable to generate new failure samples. Suppose that this occurs at timestep $\tilde{t} + 1$, and the latest batch of failure samples $\mathbf{F}^{(1)}, \ldots, \mathbf{F}^{(N)}$ were generated at timestep $\tilde{t}$. We reuse these failure samples for the remainder of the simulation. Each of these $N$ samples, besides the sensor observation, corresponds to a belief about a possible state of the intruder. We use Kalman Filters to represent these $N+1$ beliefs: $\{(\bm{\mu}^{(0)}, \bm{\Sigma}^{(0)}), (\bm{\mu}^{(1)}, \bm{\Sigma}^{(1)}), \dots, (\bm{\mu}^{(N)}, \bm{\Sigma}^{(N)})\}$.

The first filter represents our belief based on sensor observation. At timestep $\tilde{t} + 1$, we initialize our belief to be diffuse, $\bm{\mu}^{(0)}=\mathbf{o}_{\tilde{t}+1}, \bm{\Sigma}^{(0)} = 3\gamma \mathbf{I}_4$, to avoid overconfidence about the current sensor measurement. At a future timestep $t'$, given a new observation $\mathbf{o}_{t'}$ we update our belief: \begin{align*}
    \bm{\mu}_p^{(0)} &\gets \mathbf{T}_s\bm{\mu}^{(0)} + \frac{1}{2} (\mathbf{o}_{t'})_{[3:4]} \\
    \bm{\Sigma}_p^{(0)} &\gets \mathbf{T}_s \bm{\Sigma}^{(0)} \mathbf{T}_s^{\mathsf{T}} + \mathbf{\Sigma}_s \\
    \mathbf{K} &\gets \bm{\Sigma}_p^{(0)} (\bm{\Sigma}_p^{(0)} + \bm{\Sigma}_o)^{-1}\\
    \bm{\mu}^{(0)}_{[1:2]} &\gets \bm{\mu}_p^{(0)} + \mathbf{K}((\mathbf{o}_{t'})_{[1:2]} - \bm{\mu}_p^{(0)})\\
    \bm{\mu}^{(0)}_{[3:4]} &\gets (\mathbf{o}_{t'})_{[3:4]} \\
    \bm{\Sigma}^{(0)}_{1:2, 1:2} &\gets (\mathbf{I} - \mathbf{K}) \mathbf{\Sigma}_p^{(0)}, \text{where}\\
    \mathbf{T}_s &= \begin{bmatrix}
        1 & 0 & 0.5 & 0 \\
        0 & 1 & 0 & 0.5
        \end{bmatrix} \\
    \bm{\Sigma}_o &= \lambda \mathbf{I}_2, \bm{\Sigma}_s = \frac{\lambda}{4} \mathbf{I}_2 
\end{align*}

The intruder vehicle is free to perform any acceleration; therefore, we cannot model the transition dynamics of the intruder velocity, and we always represent our belief about the intruder velocity as a Gaussian with mean $\bm{\mu}^0_{[3:4]}$ equal to our current observed value and with a variance equal to the sensor noise variance. The state transitional variance $\mathbf{\Sigma}_s$ is determined based on the fact that the current observed intruder velocity is used to estimate the intruder mean velocity over the past timestep and predict the state transition, and thus sensor noise is the only source of transitional uncertainty. 

The other $N$ filters, $(\bm{\mu}^1, \bm{\Sigma}^1), \dots, (\bm{\mu}^N, \bm{\Sigma}^N)$, represent our beliefs based on the $N$ failure trajectories. They are initialized and updated in a similar fashion, except that to update $(\bm{\mu}^i, \bm{\Sigma}^i)$ at timestep $t'$ we use $\mathbf{f}_{t'}^{(i)}$ in place of $\mathbf{o}_{t'}$.

We sample $M$ possible intruder states from each filter and select ``plausible" intruder states. An intruder state is considered plausible if its positional component is within $\eta$ of the current observed intruder position. The IDM policy is used to compute an acceleration for each plausible intruder state. The most conservative (lowest-acceleration) action is performed. \Cref{robust_planning_alg} elaborates on our planning algorithm.

\begin{algorithm}[t]
\caption{Robust Planning with Failure Samplers}
\label{robust_planning_alg}
\begin{algorithmic}[1] 
\small
\Require traffic simulator $\rho_{\mathbf{S}}(\mathbf{s}_0, \bm{\epsilon}, t)$ which performs a simulation from the current timestep $t$ with the specified initial traffic state $\mathbf{s}_0$ and temporal observation error sequence $\bm{\epsilon}$, where $\mathbf{S} \in \{\text{east}, \text{west}, \text{south}, \text{north}\}$ is the scenario. It returns the state trajectory and robustness value of the resulting simulation.
\Require real-time failure samplers for each scenario, $p_{\theta_{\mathbf{S}}}(\bm{\epsilon} \mid \rho_{\text{threshold}}, \mathbf{s}_0)$ for all $\mathbf{S} \in \{\text{east}, \text{west}, \text{south}, \text{north}\}$. It takes as input the initial state $\mathbf{s}_0$ and robustness threshold $\rho_{\text{threshold}}$ and returns an observation error sequence $\bm{\epsilon}$.
\Require IDM Policy, IDM($\mathbf{s}$), which returns the acceleration to perform $a$ based on the estimated current state $\mathbf{s}$.
\Require OPTIMIZER($\mathbf{o}, \mathbf{F}^{(1) \sim (N)}, t$) which returns the optimal action sequence $a_{t+1 \sim 23}$ based on the current observation $\mathbf{o}$, the failure trajectories $\mathbf{F}^{(1) \sim (N)}$, and the current timestep $t$.
\Require UPDATE($\bm{\mu}, \bm{\Sigma}, \mathbf{o}$), which updates the parameters $\bm{\mu}, \bm{\Sigma}$ of a Kalman Filter with observation $\mathbf{o}$
\Require total failure sample size $N'$, elite failure sample size $N$, sample plausibility threshold $\eta$
    \State Initial timestep $t \gets 0$
    \State Select failure sampler $p_{\theta}$ for current scenario
    \State Observe current state $\mathbf{o}_t$ 
    \While{$\text{ego's distance to intersection} \geq 35$} \Comment{Planning Phase}
        \State generate sensor noise samples $\{\bm{\epsilon}^{(n)}\}_{n=1}^{N'} \sim p_{\theta}(\bm{\epsilon} \mid 0, \mathbf{o}_t)$
        \State simulate $\{\rho^{(n)}, \mathbf{F}^{(n)} \gets \rho_{\mathbf{S}}(\mathbf{o}_t, \bm{\epsilon}^{(n)}, t)\}_{n=1}^{N'}$
        \State select $N$ state trajectories with lowest robustness $\{\mathbf{F}^{(i)}\}_{i=1}^N$
        \State compute plan $a_{t+1 \sim 23} \gets \text{OPTIMIZER}(\mathbf{o}_t, \{\mathbf{F}^{(i)}\}_{i=1}^N, t)$
        \State perform action $a_{t+1}$
        \State $t \gets t+1$; make new observation $\mathbf{o}_t$
    \EndWhile

    \State initialize filter for sensor observation $\bm{\mu}^0 \gets \mathbf{o}_t$; $\bm{\Sigma}^0 \gets 3\gamma \mathbf{I}_4$
    \State initialize filters for failure samples $\{\bm{\mu}^i \gets \mathbf{f}_t^{(i)}, \bm{\Sigma}^i \gets 3\gamma \mathbf{I}_4\}_{i=1}^N$
    \While{$t < 23$} \Comment{Policy-Based Phase}
        \State $\bm{\mu}^0, \bm{\Sigma}^0 \gets \text{UPDATE}(\bm{\mu}^0, \bm{\Sigma}^0, \mathbf{o}_t)$
        \State $\{\bm{\mu}^i, \bm{\Sigma}^i \gets \text{UPDATE}(\bm{\mu}^i, \bm{\Sigma}^i, \mathbf{f}_t^{(i)})\}_{i=1}^N$
        \State sample possible intruder states $\{\{\mathbf{s}^i_j\}_{j=1}^M \sim \mathcal{N}(\bm{\mu}^i, \bm{\Sigma}^i)\}_{i=0}^N$
        \State plausible states $\mathcal{S} \gets \{\mathbf{s}_j^i \mid \|(\mathbf{s}_j^i)_{[1:2]} - (\mathbf{o}_t)_{1:2}\|_2 \leq \eta\}$
        \State perform action $a \gets \min_{\mathbf{s} \in \mathcal{S}} \text{IDM}(\mathbf{s})$
        \State $t \gets t+1$; make new observation $\mathbf{o}_t$
    \EndWhile
\end{algorithmic}
\end{algorithm}

\end{document}